\pdfoutput=1

\documentclass[11pt]{article}

\usepackage{ACL2023}

\usepackage{times}
\usepackage{latexsym}
\usepackage{graphicx}
\usepackage{amsfonts}
\usepackage{booktabs}
\usepackage{multirow}
\usepackage{makecell}
\usepackage{amsmath}
\usepackage{subfigure}

\usepackage[T1]{fontenc}

\usepackage[utf8]{inputenc}

\usepackage{microtype}

\usepackage{inconsolata}

\title{Neuron-Level Knowledge Attribution in Large Language Models}

\author{Zeping Yu \quad Sophia Ananiadou\\
  Department of Computer Science, National Centre for Text Mining \\
  The University of Manchester  \\
  \texttt{\{zeping.yu@postgrad. sophia.ananiadou@\}manchester.ac.uk}}

\begin{document}
\maketitle
\begin{abstract}
Identifying important neurons for final predictions is essential for understanding the mechanisms of large language models. Due to computational constraints, current attribution techniques struggle to operate at neuron level. In this paper, we propose a static method for pinpointing significant neurons. Compared to seven other methods, our approach demonstrates superior performance across three metrics. Additionally, since most static methods typically only identify "value neurons" directly contributing to the final prediction, we propose a method for identifying "query neurons" which activate these "value neurons". Finally, we apply our methods to analyze six types of knowledge across both attention and feed-forward network (FFN) layers. Our method and analysis are helpful for understanding the mechanisms of knowledge storage and set the stage for future research in knowledge editing. The code is available on \url{https://github.com/zepingyu0512/neuron-attribution}.
\end{abstract}

\section{Introduction}
Transformer-based large language models (LLMs) \cite{brown2020language,ouyang2022training,chowdhery2023palm} possess remarkable capabilities for storing factual knowledge, which is important for downstream tasks including question answering \cite{jiang2021can} and reasoning \cite{rajani2019explain}. While recent studies \cite{dai2021knowledge,meng2022locating,geva2023dissecting,yu2023characterizing,chen2024journey} have made significant progress in understanding knowledge localization and the information flow from inputs to predictions, it is still hard to identify exact parameters for knowledge storage in LLMs due to several reasons. Firstly, existing studies often depend on causal tracing \cite{pearl2001direct,vig2020investigating} and integrated gradients \cite{sundararajan2017axiomatic} for knowledge attribution. However, many studies \cite{stolfo2023mechanistic,zhao2024explainability,wu2024usable} point out that the computational complexity of forward and backward operations in these methods restricts their applicability to millions of neurons in LLMs, which are proved as fundamental units for storing knowledge \cite{geva2020transformer,dai2021knowledge,geva2022transformer,nanda2023fact}. Secondly, while a few studies \cite{dar2022analyzing,geva2022transformer} have devised methods for analyzing neurons, they often lack comparisons with other methods. Therefore, how to identify important neurons in LLMs is still unclear. Lastly, existing methods typically concentrate on either attention or feed-forward network (FFN) module, often lacking evaluation of the other module. It is crucial to quantitatively compare the importance of both attention and FFN layers.

\begin{figure}[htb]
  \centering
  \includegraphics[width=0.85\columnwidth]{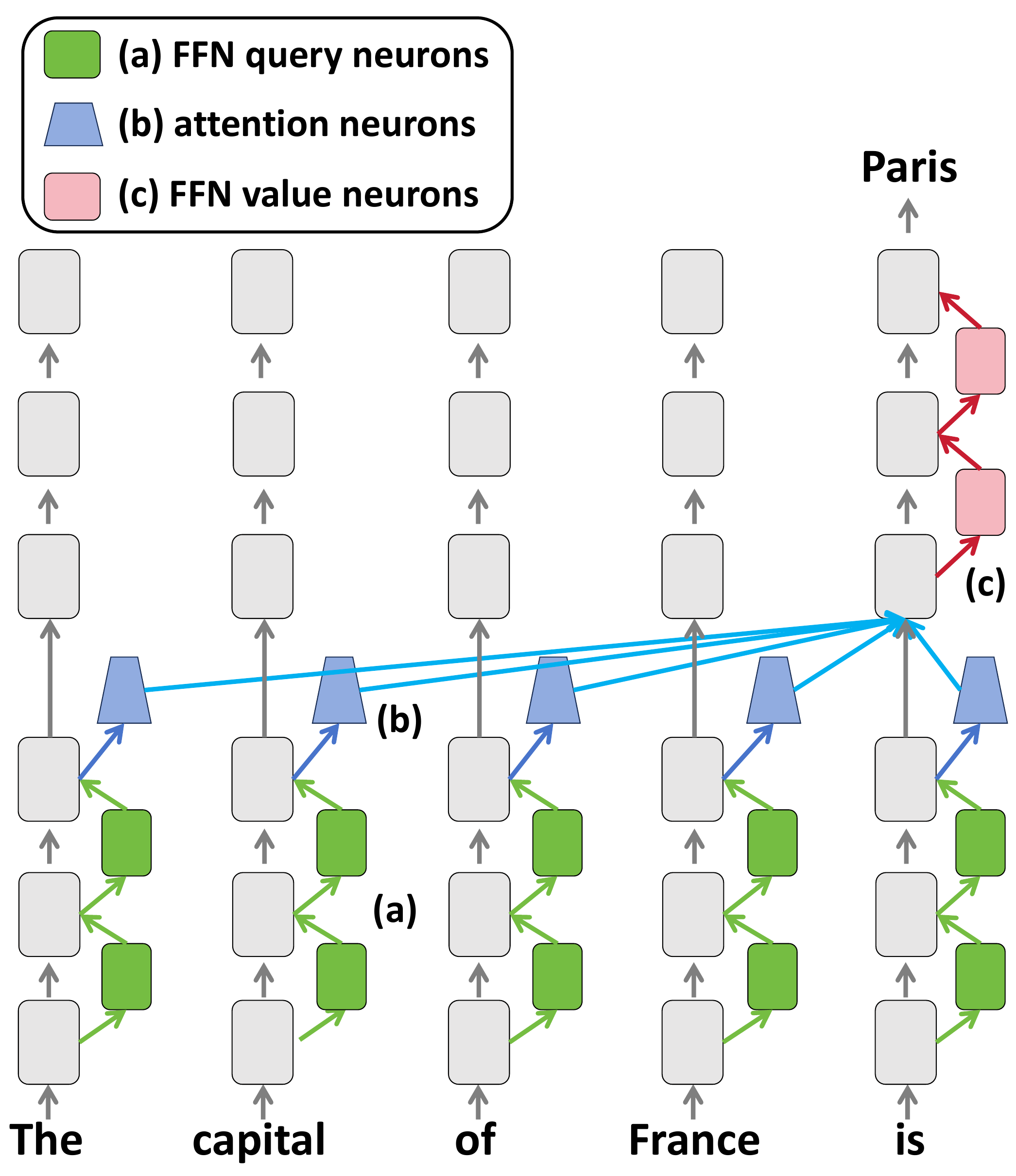}
  \caption{(a) Query neurons in shallow FFN layers. (b) Attention query/value neurons in attention heads. (c) Value neurons in deep FFN layers.}
\vspace{-10pt}
\end{figure}

In this paper, we focus on neuron-level attribution methods. We analyze the distribution change caused by each neuron and discover that both the neuron's coefficient score and the final prediction's ranking, when projecting this neuron's subvalue into vocabulary space, play significant roles. Based on this finding, we employ log probability increase as importance score, enabling the identification of neurons that contribute significantly to final predictions. Compared with seven other static methods, our proposed method achieves the best performance on three metrics. Furthermore, since the identified neurons directly contribute to the final predictions' probability, we also develop a static method to identify "query neurons" that aid in activating these "value neurons". Specifically, we calculate the inner products between the query neurons and value neurons as importance scores. 

Based on our proposed methods, we analyze six types of knowledge in both attention and FFN layers, yielding numerous valuable insights (Figure 1): 1) Both attention and FFN layers can store knowledge, and all important neurons directly contribute to knowledge prediction are in deep layers. 2) In attention layers, knowledge with similar semantics (e.g. language, country, city) tends to be stored in the same heads. Knowledge with distinct semantics (e.g. country, color) is stored in different heads. 3) While numerous neurons contribute to the final prediction, intervening on a few value neurons (300) or query neurons (1000) can significantly influence the final prediction. 4) FFN value neurons are mainly activated by medium-deep attention value neurons, while these attention neurons are mainly activated by shallow/medium FFN query neurons.

Overall, our contributions are as follows:

a) We design a static method for neuron-level knowledge attribution in large language models. Compared with seven static methods, our method achieves the best performance under three metrics.

b) As the identified neurons usually directly contribute to the final predictions, we design a static method to identify the "query neurons" activating these "value neurons".

c) We analyze the localization of six types of knowledge in both attention and FFN layers. Our analysis is helpful for understanding the mechanisms of knowledge storage in language models. 

\section{Related Work}
\subsection{Attribution Methods for Transformers}
Determining how to attribute the important parameters for final predictions is a crucial question. Gradient-based methods \cite{sundararajan2017axiomatic,kindermans2019reliability,miglani2020investigating,lundstrom2022rigorous} and causal tracing methods \cite{pearl2001direct,vig2020investigating,meng2022locating,goldowsky2023localizing,zhang2023towards,wu2024interpretability,hase2024does} are widely utilized for this purpose. The core idea is calculating how much an internal module affects the final predictions, requiring multiple forward and/or backward operations \cite{wu2024usable}. Due to the computational overhead, these methods are usually applied on hidden states \cite{meng2022locating,geva2023dissecting,stolfo2023mechanistic}, rather than neurons. Another type of studies tend to require only one forward pass for each sentence, typically relying on saliency scores such as attention weights \cite{vig2019bertviz,jaunet2021visqa,yeh2023attentionviz,wang2023label,li2023towards} and FFN neurons' coefficient scores \cite{geva2022transformer,lee2024mechanistic}. However, the validity of attributions is challenged by many studies \cite{serrano2019attention,jain2019attention,wiegreffe2019attention,mohankumar2020towards,ethayarajh2021attention,bai2021attentions}. Lack of evaluation methods results in an ongoing debate about the faithfulness of saliency score methods.

\subsection{Mechanistic Interpretability}
Mechanistic interpretility \cite{Chris2022,nanda2023progress} aims to reverse engineer the circuits from inputs to the final prediction. An essential technology involves projecting internal vectors into the vocabulary space, where numerous studies have discovered interpretable results \cite{nostalgebraist2020,geva2020transformer,geva2022transformer,dar2022analyzing,pal2023future}. Most studies focus on analyzing the attention heads' roles in different cases and tasks \cite{elhage2021mathematical,olsson2022context,wang2022interpretability,hanna2023does,lieberum2023does,conmy2023towards,gould2023successor}. Also, superposition \cite{elhage2022toy,nanda2023fact,gurnee2023finding} and dictionary learning \cite{bricken2023monosemanticity,he2024dictionary} are important for understanding neurons in transformers. 

\section{Methodology}
In this section, we aim to locate important neurons for specific predictions. We introduce the background in Section 3.1, and analyze the distribution change caused by neurons in Section 3.2. Based on the analysis, we introduce our proposed method for locating the "value neurons" that contribute to the final predictions directly in Section 3.3, and propose a static method to locate the "query neurons" that activate these "value neurons" in Section 3.4.

\subsection{Background}
First, we introduce the inference pass from inputs to the final prediction. Given an input sentence $X=[t_1, t_2, ..., t_T]$ with $T$ tokens, the model generates the next token's probability distribution $y$ over $B$ tokens in vocabulary $V$. The embedding matrix $E \in \mathbb{R}^{B \times d}$ transforms each $t_i$ at position $i$ into a word embedding $h_i^0 \in \mathbb{R}^{d}$. Then the word embeddings are transformed by $L+1$ transformer layers ($0th-Lth$), each has a multi-head self-attention layer (MHSA) and a FFN layer. The layer output $h_i^l$ (position $i$, layer $l$) is the sum of the layer input $h_i^{l-1}$ (previous layer's output), the attention output $A_i^l$, and the FFN output $F_i^l$:
\begin{equation}
h_i^l = h_i^{l-1} + A_i^{l} + F_i^{l}
\end{equation}
Finally, the last position's $Lth$ layer output is used to compute the final probability distribution $y$ by multiplying the unembedded matrix $E_u \in \mathbb{R}^{B \times d}$:
\begin{equation}
y = softmax(E_u \, h_T^{L})
\end{equation}
Specifically, the attention layer's output is computed by a weighted sum over $H$ heads on $T$ positions, and the FFN layer's output is computed by a nonlinear function $\sigma$ on two linear transformations. 
\begin{equation}
A_i^l = \sum_{j=1}^H ATTN_j^l(h_1^{l-1}, h_2^{l-1}..., h_T^{l-1})
\end{equation}
\begin{equation}
F_i^l = W_{fc2}^l\sigma (W_{fc1}^l (h_i^{l-1}+A_i^l))
\end{equation}
where $W_{fc1}^l \in \mathbb{R}^{N \times d}$ and $W_{fc2}^l \in \mathbb{R}^{d \times N}$ are two matrices. \citet{geva2020transformer} find FFN output can be represented as a weighted sum of FFN neurons:
\begin{equation}
F_i^l = \sum_{k=1}^N {m_{i,k}^l fc2_{k}^l}
\end{equation}
\begin{equation}
m_{i,k}^l = \sigma (fc1_k^l \cdot (h_i^{l-1}+A_i^l))
\end{equation}
The FFN output $F_i^l$ is computed by a weighted sum of $fc2$ vectors. $fc2_k^l$ is the $kth$ column of $W_{fc2}^l$ (named FFN subvalue), and its coefficient score $m_{i,k}^l$ is computed by non-linear $\sigma$ on the inner product between the residual output $h_i^{l-1}+A_i^l$ and $fc1_k^l$ (named FFN subkey), the $kth$ row of $W_{fc1}^l$. Similarly, the attention output $A_i^l$ can be represented as a sum of head outputs, each being a weighted sum of value-output vectors on all positions:
\begin{equation}
A_i^l = \sum_{j=1}^H \sum_{p=1}^T \alpha_{i,j,p}^l W^o_{j,l} (W^v_{j,l} h_p^{l-1})
\end{equation}
\begin{equation}
\alpha_{i,j,p}^l = softmax(W^q_{j,l} h_i^{l-1} \cdot W^k_{j,l} h_p^{l-1})
\end{equation}
where $W_{j,l}^q, W_{j,l}^k, W_{j,l}^v, W_{j,l}^o \in \mathbb{R}^{d \times {d/H}}$ are the query, key, value and output matrices of the $jth$ head in the $lth$ layer. The query and key matrices compute the attention weight $\alpha_{i,j,p}^l$ on the $pth$ position, then calculate the softmax function across all positions. The value and output matrices transform the $pth$ position input vector into the $pth$ value-output vector. Each head output is the weighted sum of value-output vectors on all positions.

\paragraph{Definition of "neuron".} As discussed in Eq.5-6, the $kth$ FFN neuron is the $kth$ subvalue $fc2_k^l$, whose coefficient score $m_{i,k}^l$ is calculated by its corresponding subkey $fc1_k^l$. In Eq.7, each attention output can be represented as a direct addition of $T \times H$ vectors, when taking the position value-output vector $W^o_{j,l} (W^v_{j,l} h_p^{l-1})$ as fundamental units. Moreover, the position value-output vector can also be regarded as a weighted sum of attention neurons. Similar to the definition of FFN neurons, we regard the $kth$ column of $W^o_{j,l}$ in Eq.7 as the $kth$ attention subvalue in this head, whose subkey is the $kth$ row of $W^v_{j,l}$. When taking the attention neurons as fundamental units, the final output is the sum of $L \times (T \times H \times d/H + N) + 1$ vectors. 

\subsection{Distribution Change Caused by Neurons}
The final vector $h_T^L$ has important information for predicting the final token. As it is computed by a direct sum of various neuron-level vectors, the relevant information for making the final prediction must be stored in one or many neurons. The final vector $h_T^l$ can be regarded as the sum of one neuron $v$ and another vector $x=h_T^l-v$. We consider the probability change $p(w|x+v)-p(w|x)$ caused by $v$ for prediction token $w$. 
We aim to explore which components of $v$ are significant for amplifying the probability change. This allows us to develop static methods for locating crucial neurons.

As the probability change is nonlinear, analyzing the exact contribution of neuron $v$ is challenging. For a more concise analysis, we term the score $e_w \cdot x$ vector $x$'s bs-value (before-softmax value) on token $w$, where $e_w$ is the $wth$ row of the unembedded matrix $E_u$. A token's bs-value directly corresponds to the probability of this token. Bs-values of all vocabulary tokens on vector $x$ are:
\begin{equation}
    bs(x) = [bs^x_1, bs^x_2, ..., bs^x_w, ..., bs^x_B]
\end{equation}
For vector $x$, if $bs^x_w$ is the largest among all the bs-values, the probability of word $w$ will also be the highest. The probability of each token for $x$ and $x+v$ can be computed by all the bs-values:
\begin{equation}
p (w|x) = \frac{exp(bs^x_w)}{exp(bs^x_1) + ... + exp(bs^x_B) }
\end{equation}
\begin{equation}
p (w|x+v) = \frac{exp(bs^{x+v}_w)}{exp(bs^{x+v}_1) + ... + exp(bs^{x+v}_B) }
\end{equation}
where bs-value $bs^{x+v}_w$ is equal to $bs^x_w + bs^v_w$:
\begin{equation}
bs(x+v) = bs(x)+bs(v)
\end{equation}

Although the probability change is nonlinear, the change on each token's bs-value is linear. In order to analyze which components of $v$ is important, we design several $bs(x)$ and $bs(v)$ and compute the distribution change. Assume there are four tokens in vocabulary space, and $bs(x)=[1,2,3,4]$. The probability distribution of $x$ is $[0.03, 0.09, 0.24, 0.64]$. We design several $v$ and compute the probability distribution of $p(x+v)$. The details are shown in Table 1.

\begin{table}[htb]
    \centering
    \scalebox{0.925}{\begin{tabular}{ccc}
        \toprule
        $bs(v)$ & $bs(x+v)$ & $p(x+v)$ \\
        \midrule
        $[1,1,1,3]$ & $[2,3,4,7]$ & $[0.01, 0.02, 0.05, 0.93]$ \\
        \midrule
        $[3,1,1,1]$ & $[4,3,4,5]$ & $[0.20, 0.07, 0.20, 0.53]$ \\
        \midrule
        $[6,4,4,4]$ & $[7,6,7,8]$ & $[0.20, 0.07, 0.20, 0.53]$ \\
        \midrule
        $[6,2,2,2]$ & $[7,4,5,6]$ & $[0.64, 0.03, 0.09, 0.23]$ \\
        \midrule
        $-[6,2,2,2]$ & $[-5,0,1,2]$ & $[0.00, 0.09, 0.24, 0.67]$ \\
        \bottomrule
    \end{tabular}}
    \caption{Probability distribution of $p(x+v)$.}
\vspace{-10pt}
\end{table}

Existing studies \cite{geva2022transformer,lee2024mechanistic} state that $p(w|x+v) \propto exp(e_w \cdot v)$. However, based on the examples provided in Table 1, not only the bs-value of token $w$, but also the bs-values of all the tokens affect the probability. For example, $bs(v)=[3,1,1,1]$ and $[6,4,4,4]$ result in the same distribution, although the bs-value of each token is enlarged.

An intuitive observation is that $v$ aids in magnifying the token with the largest bs-value. For instance, $[1,1,1,3]$ is conducive to increasing the probability of the last token, and $[3,1,1,1]$ can amplify the probability of the first token. This observation may elucidate why many neurons exhibit human-interpretable concepts when projected into the vocabulary space. Given that the vocabulary size $B$ is typically large (often exceeding 30,000), probabilities of tokens with the largest bs-values are likely to be augmented.

Another significant finding is that both the coefficient score and the neuron's bs-values play substantial roles. Compared with $[3,1,1,1]$, $[6,2,2,2]$ can can both amplify and diminish the probability change of $[3,1,1,1]$. The probability increase of the first token is magnified, while the decrease in probability of the last token is more pronounced. When the coefficient score's sign is changed (e.g. $-[6,2,2,2]$), the effect on the first token's probability changes from increasing to decreasing.

\subsection{Importance Score for "Value Neurons"}
Based on the analysis in Section 3.2, an intuitive importance score of a neuron $mv$ is $|m| \times |1/rank(w)|$, where $m$ is the coefficient score and $rank(w)$ denotes the ranking of the final token when projecting $v$ into vocabulary space. Another intuitive importance score is calculating the probability $p(w|mv)$ on token $w$. If these scores are large, $v$ will contain much information of $w$. 

However, these methods have two potential problems. On one hand, they only consider the effect of $v$, overlooking the varying importance of $v$ under different $x$ conditions. On the other hand, we usually hope to analyze the importance of different modules' combination. Therefore, it is better that the importance score $Imp$ satisfies $Imp(x+v) \approx Imp(x) + Imp(v)$.

To address these problems, we design log probability increase as importance score for both layer-level and neuron-level vectors. If $v^l$ is a vector in $lth$ attention layer, the importance score of $v^l$ is:
\begin{equation}
Imp(v^l) = log(p(w|v^l+h^{l-1})) - log(p(w|h^{l-1}))
\end{equation}
where the probability of each vector is computed by multiplying the vector with $E_u$ (see Eq.2). If $v^l$ is a vector in $lth$ FFN layer, we compute the importance score by replacing $h^{l-1}$ as $h^{l-1}+A^{l}$ in Eq.13. In Eq.13, $v^l$ is not the only element controlling the importance score. Also, it is convenient for analyzing the combination of different modules.

\subsection{Importance Score for "Query Neurons"}
As discussed in Section 3.3, the proposed attribution methods can effectively identify the "value neurons" containing crucial information for the final prediction. However, in addition to these "value neurons", there exist "query neurons" that aid in activating these neurons, even if they may not directly contain information about $w$. In this section, we propose a static method to identify these "query neurons" based on Eq.1, Eq.5, and Eq.6. Since the $fc2$ vectors do not change, the coefficient scores are the only varying element in different cases. For each "value neuron", we can compute the inner product between its subkey (see Eq.6) and each neuron/subvector within the residual output (see Eq.1). Despite the presence of a nonlinear function $\sigma$ for computing the coefficient score, it usually does not affect the relative value between different neurons/subvectors. Therefore, if a "query" neuron/subvector exhibits a larger inner product with the subkey compared to another one, it is more helpful for activating the "value neuron".

\section{Experiments}
In this section, we compare our neuron-level attribution method with seven other methods in Section 4.1. Then we analyze the localization of six types of knowledge using our method in Section 4.2.

\subsection{Comparison of Attribution Methods}
We compare the proposed method in Eq.13 with seven other methods. For each sentence, we apply every method to identify top10 FFN neurons, and evaluate the attributed neurons using three metrics.

\paragraph{Dataset.} We extract query-answer pairs with six types of answer tokens (language, capital, country, color, number, month) from TriviaQA \cite{joshi2017triviaqa}. To explore the mechanism of knowledge storage, we extract all the sentences where the correct token ranks within the top10 predictions and higher than other words in the same knowledge in GPT2-large \cite{radford2019language} and Llama-7B \cite{touvron2023llama}. There are 1,350 sentences for GPT2-large and 3,141 sentences for Llama-7B. 

\paragraph{Models.} To compare the differences between large and small models in terms of knowledge storage, we conduct experiments on Llama-7B and GPT2-large. Llama-7B consists of 32 layers, with each attention layer comprising 32 heads, each head containing 128 neurons and each FFN layer containing 11,008 neurons. GPT2-large has 36 layers with 20 heads per attention layer, 64 neurons per head, and 5,120 neurons per FFN layer.

\paragraph{Attribution methods.} We compare our method with seven static methods. We use each method to attribute the FFN neurons with top10 scores for the correct knowledge token $w$. Similar to Eq.5, each neuron $mv$ is the product of the coefficient score $m$ and $fc2$ vector $v^l$. Here are the methods:

\noindent a) (proposed method) log probability increase: $log(p(w|mv^l+A^l+h^{l-1}))-log(p(w|A^l+h^{l-1}))$

\noindent b) log probability: $log(p(w|mv^l))$, attributing the same neurons with $p(w|mv^l)$. This is similar to direct logit attribution (DLA) in \citet{wang2022interpretability}.

\noindent c) probability increase: $p(w|mv^l+A^l+h^{l-1})-p(w|A^l+h^{l-1})$

\noindent d) norm: $|v^l|$

\noindent e) coefficient score: $|m|$

\noindent f) ranking in vocabulary space: $1/rank(w)$

\noindent g) $|m| \times |v^l|$, introduced in \citet{geva2022transformer}.

\noindent h) $|m| \times 1/rank(w)$

\paragraph{Metrics.} We devise three metrics to evaluate the attribution methods. After attributing the top10 FFN neurons by each method, we intervene on these neurons by setting the top10 neurons' parameters to zero. Subsequently, we rerun the model and compute the Mean Reciprocal Rank (MRR) score of the correct token $w$, the probability of $w$ (prob), and the log probability of $w$ (logp). An attribution method is considered superior when it exhibits greater decreases in these metrics.

\begin{table}[htb]
  \centering
  \scalebox{0.95}{\begin{tabular}{ccccccc}
    \toprule
   \multicolumn{1}{c}{} & \multicolumn{3}{c}{GPT2-large} & \multicolumn{3}{c}{Llama-7B}\\ 
        & MRR & prob & logp & MRR & prob & logp \\
    \midrule
    o) & 0.361 & 7.1 & -3.15 & 0.551 & 21.5 & -2.24 \\
    a) & \textbf{0.201} & \textbf{3.4} & \textbf{-4.06} & \textbf{0.312} & \textbf{9.2} & \textbf{-3.91} \\
    b) & 0.214 & 3.6 & -3.91 & 0.339 & 10.8 & -3.35 \\
    c) & 0.219 & 3.7 & -3.92 & 0.345 & 10.0 & -3.57 \\
    d) & 0.363 & 7.1 & -3.14 & 0.549 & 21.3 & -2.25 \\
    e) & 0.439 & 8.6 & -3.10 & 0.529 & 22.9 & -2.35 \\
    f) & 0.306 & 5.8 & -3.40 & 0.493 & 18.1 & -2.49 \\
    g) & 0.394 & 8.1 & -3.06 & 0.523 & 22.6 & -2.39 \\
    h) & 0.232 & 4.0 & -3.80 & 0.389 & 13.0 & -3.06 \\
    \bottomrule
  \end{tabular}}
  \caption{Results of attribution methods on two models.}
\vspace{-10pt}
\end{table}

\paragraph{Results and analysis.} The results of the original model (first line) and eight attribution methods are shown in Table 1. In comparison with the other seven methods, our attribution method (second line) attributes more important neurons, resulting in the most significant reduction across all metrics in both GPT2 and Llama. Specifically, when only intervening ten FFN neurons, the probability of the correct knowledge token reduces from 7.1\% to 3.4\% in GPT2, and from 21.5\% to 9.2\% in Llama-7B. This indicates that there are several neurons storing much important information for knowledge storage, and our method can locate these neurons.

The attribution methods of norm $v^l$ (d) and $m \times |v^l|$ (g) are not useful, which indicates the norm of neurons is not important for attribution. Using $|m| \times 1/rank(w)$ (h) has good results, which is better than $1/rank(w)$ (f). The ranking of tokens in vocabulary space for projected neurons is a good saliency score, and the coefficient score can enlarge the distribution change, aligning our analysis in Section 3.2. Only using coefficient score (e) is not helpful for attribution. The role of coefficient score is to enhance the probability change caused by the neuron, but whether the neuron is useful for the selected token depends on the neuron itself. There are other tokens competing with the correct knowledge token, so the neurons with large coefficient scores may be related to these tokens.

\begin{figure}[htb]
  \centering
  \includegraphics[width=0.92\columnwidth]{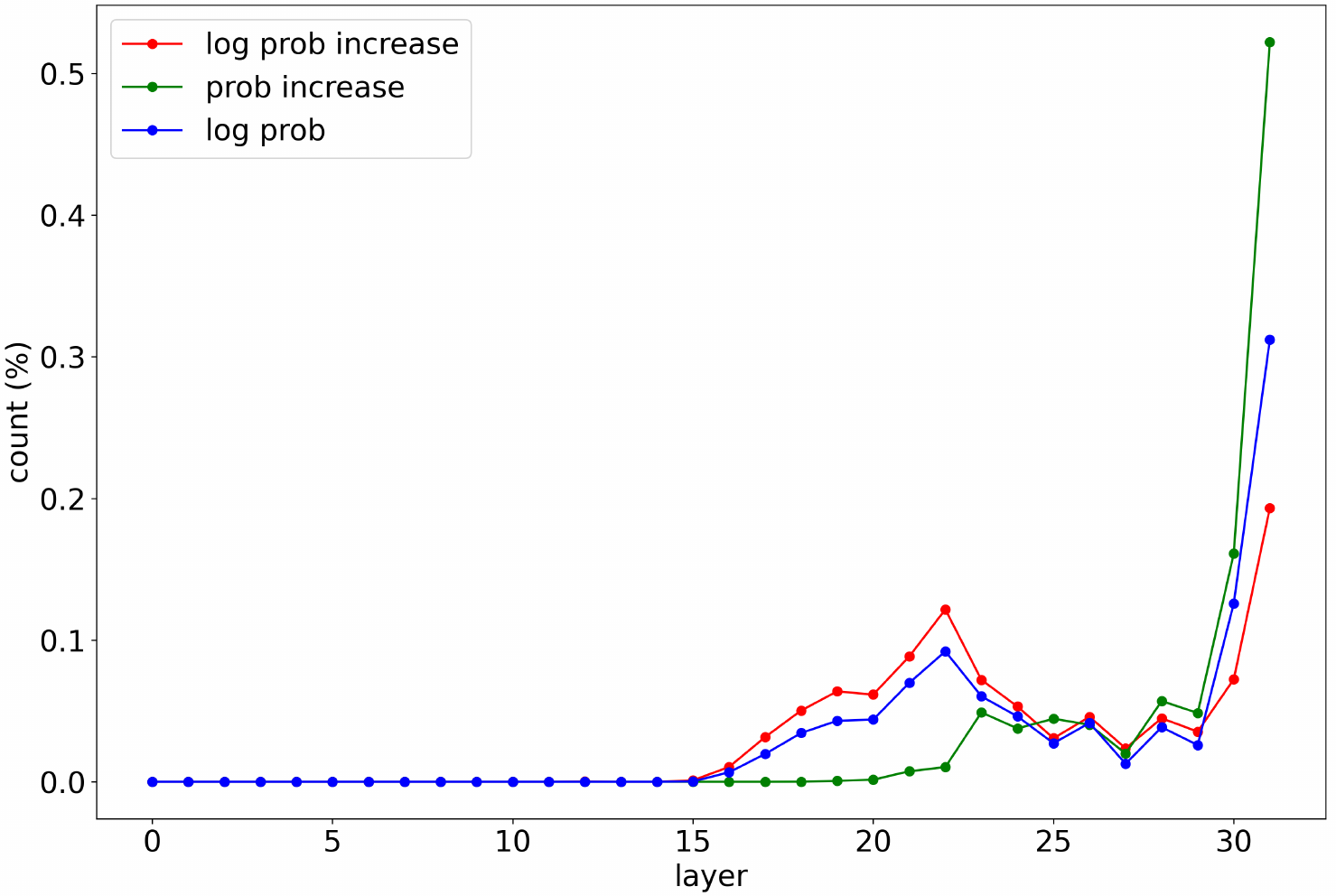}
  \caption{Neuron distribution on all layers in Llama-7B.}
\vspace{-10pt}
\end{figure}

Compared to log probability (b), employing log probability increase (a) can attribute more important neurons. This aligns with the analysis in Section 3.2 and 3.3: not only neuron $v$, but also $x$ affects $p(w|x+v)-p(w|x)$. Compared with probability increase (c), log probability increase achieves better results. We analyze the distribution of neurons across all layers in Llama attributed by log probability increase, log probability, and probability increase, as depicted in Figure 2. GPT2 has similar results, detailed in Appendix A. The neurons attributed by probability increase are on deepest layers ($23th-31th$), while other two methods can attribute neurons among $17th$ to $31th$ layers.

To delve into the reason of this phenomenon, we analyze the difference of importance score when adding the same vector $v$ on different $x$. As discussed in Eq.13, the importance score of $v$ is computed by $log(p(w|x+v))-log(p(w|x))$. Therefore, the importance score is related to the curve of $F(a)=log(p(w|a))$. To analyze this curve, we compute the final vector $h_T^L$ and the $0th$ layer input vector $h_T^0$ on each sentence, and divide $h_T^L-h_T^0$ into 61 segments, where each segment is $Seg_S = h_T^0+S(h_T^L-h_T^0)/60$ ($S$ is the segment index from 0 to 60). Then we compute the probability $p(w|Seg_S)$ and log probability $log(p(w|Seg_S))$ at each segment index for every sentence. The average score on Llama-7B is shown in Figure 3.

\begin{figure}[thb]
	\centering
	\subfigure{
		\begin{minipage}{0.47\textwidth}
			\includegraphics[width=0.47\textwidth]{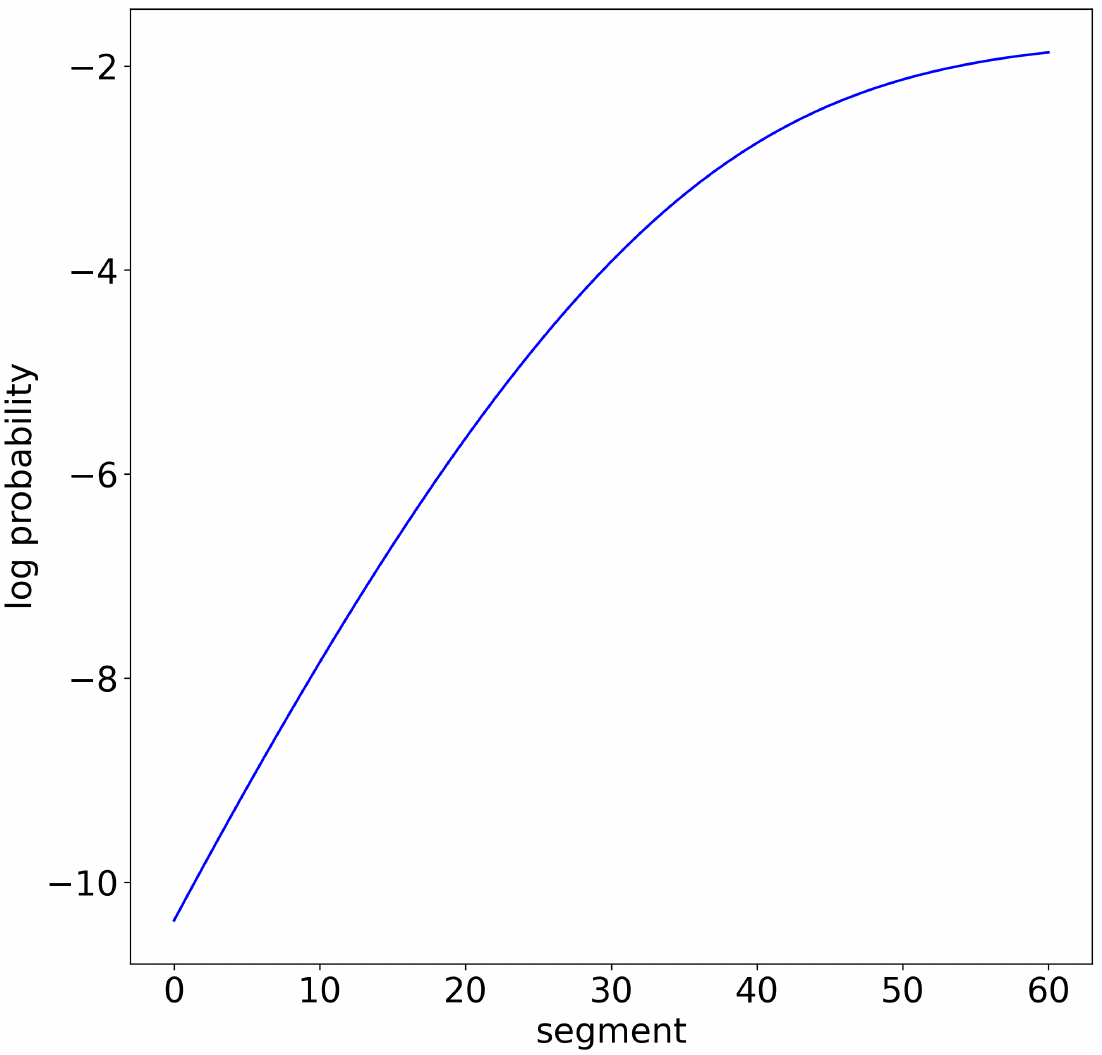}
                \includegraphics[width=0.48\textwidth]{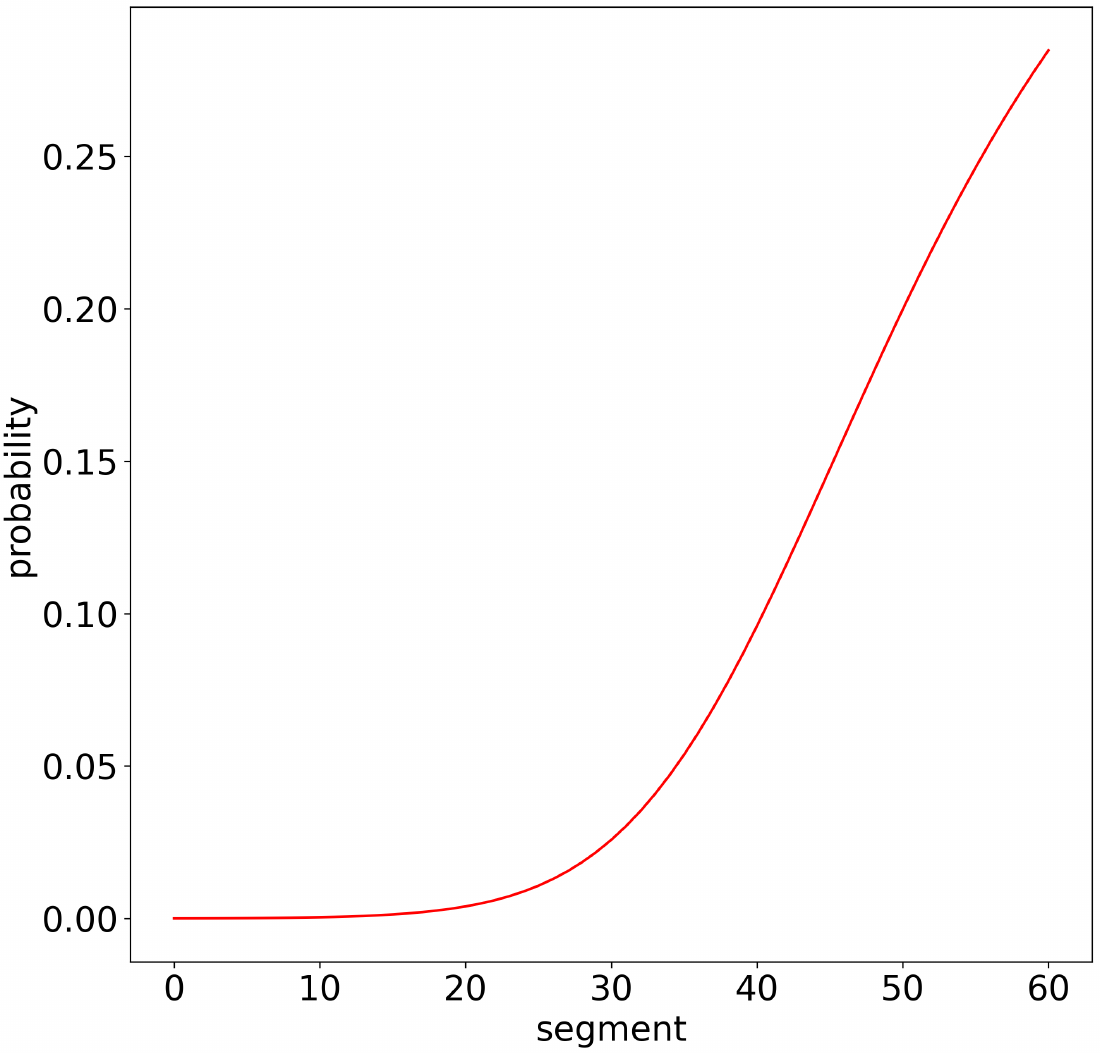}
		\end{minipage}
	}
	\caption{Curves of log probability increase (left) and probability increase (right) on Llama-7B.}
\vspace{-10pt}
\end{figure}

The curve of log probability increase exhibits an approximately linear shape from 0 to 40 segments, while the curve of probability increase shows a linear trend from 40 to 60 segments. This observation elucidates the findings in Figure 2: employing probability increase is more inclined to attribute neurons in the deepest layers, whereas log probability increase tends to attribute neurons in medium-deep layers. Despite the slower slope of the log probability increase curve in very deep layers, it still effectively attributes neurons in very deep layers (as depicted in Figure 2). This maybe because neurons in very deep layers contain substantial information, and even when the importance score decreases, it remains relatively large. In later sections, we use log probability increase as importance score for exploration, as this method can identify the important neurons in both medium-deep layers and very deep layers, and its experimental results are the best. Nevertheless, reproducing the importance of the deepest layers may be a prospective avenue for developing improved attribution methods.

\subsection{Exploration on Different Knowledge}
We take log probability increase as importance score, and analyze six types of knowledge: language (lang), capital (capi), country (cnty), color (col), number (num), and month (mon). We evaluate the knowledge storage in attention and FFN layers at layer-level, head-level, and neuron-level. We also compute the storage of query layers and query neurons.

\begin{table}[htb]
  \centering
  \scalebox{0.95}{\begin{tabular}{ccccccc}
    \toprule
        & lang & capi & cnty & col & num  & mon \\
    \midrule
    G-A & 7.46 & 6.61 & 6.62 & 4.10 & 2.75  & 4.82 \\
    G-F & 0.63 & 1.89 & 1.67 & 3.51 & 4.70  & 2.81 \\
    \midrule
    L-A & 6.28 & 5.03 & 6.42 & 3.81 & 2.22  & 5.10 \\
    L-F & 1.74 & 4.30 & 2.90 & 4.02 & 5.60  & 3.05 \\
    \bottomrule
  \end{tabular}}
  \caption{Contribution of attention and FFN layers.}
\vspace{-10pt}
\end{table}

\paragraph{Layer-level knowledge storage.} We compute the sum of importance score of each attention and FFN layer in GPT2 (G-A, G-F) and Llama (L-A, L-F), shown in Table 3. Also, we calculate the top10 layer for each knowledge in Table 4, where $a_l$ and $f_l$ means $lth$ attention and FFN layer. For a clearer display, we illustrate the importance (darker color means larger importance) of top10 layers in Figure 4 (GPT2) and Figure 5 (Llama).

Both attention and FFN layers have ability to store knowledge, and all the top10 layers are in deep layers.  Information with analogous semantics (e.g., language, capital, country) tends to be stored within similar layers/modules. For instance, $a_{26}$, $a_{30}$, $a_{28}$, and $a_{22}$ in GPT2 ranks top for language, capital and country, and $a_{23}$ in Llama-7B ranks the first for these knowledge. Data with dissimilar semantics (e.g., language, color, month) typically resides in distinct layers/modules. 

\begin{table}[htb]
    \centering
    \scalebox{0.925}{\begin{tabular}{cc}
        \toprule
         & top10 important layers\\
        \midrule
        lang & $a_{26}, a_{30}, a_{32}, a_{22}, a_{31}, a_{28}, a_{23}, a_{27}, a_{19}, a_{23}$ \\
        capi & $a_{26}, a_{28}, a_{30}, a_{25}, a_{22}, f_{26}, f_{28}, a_{19}, f_{27}, f_{30}$ \\  
        cnty & $a_{26}, a_{30}, a_{28}, a_{22}, f_{29}, a_{31}, f_{26}, a_{32}, a_{25}, a_{19}$ \\  
        col & $a_{32}, f_{32}, a_{33}, f_{29}, f_{31}, a_{31}, a_{26}, f_{33}, f_{28}, a_{22}$ \\
        num & $f_{29}, f_{23}, f_{27}, f_{30}, f_{31}, f_{26}, f_{32}, a_{23}, a_{22}, f_{28}$ \\
        mon & $a_{27}, a_{26}, f_{26}, a_{25}, f_{30}, a_{28}, a_{24}, a_{22}, a_{30}, f_{27}$ \\
        \midrule
        lang & $a_{23}, a_{21}, f_{21}, a_{19}, a_{18}, a_{31}, a_{25}, a_{16}, f_{20}, f_{19}$ \\
        capi & $a_{23}, f_{21}, f_{22}, a_{18}, a_{25}, a_{21}, f_{19}, f_{20}, a_{16}, f_{24}$ \\  
        cnty & $a_{23}, a_{21}, a_{25}, f_{22}, a_{18}, a_{19}, a_{16}, f_{21}, f_{31}, a_{31}$ \\ 
        col & $f_{29}, a_{20}, f_{22}, f_{20}, a_{19}, a_{28}, a_{16}, a_{29}, a_{18}, f_{28}$ \\
        num & $f_{31}, f_{26}, f_{29}, f_{27}, a_{26}, f_{23}, f_{24}, a_{28}, f_{17}, f_{30}$ \\
        mon & $a_{21}, a_{19}, f_{19}, a_{16}, f_{31}, a_{23}, a_{28}, f_{30}, f_{17}, f_{18}$ \\
        \bottomrule
    \end{tabular}}
    \caption{Top10 important layers in GPT2 (first block) and Llama (second block).}
\vspace{-10pt}
\end{table}

\begin{figure}[htb]
  \centering
    \includegraphics[width=0.98\columnwidth]{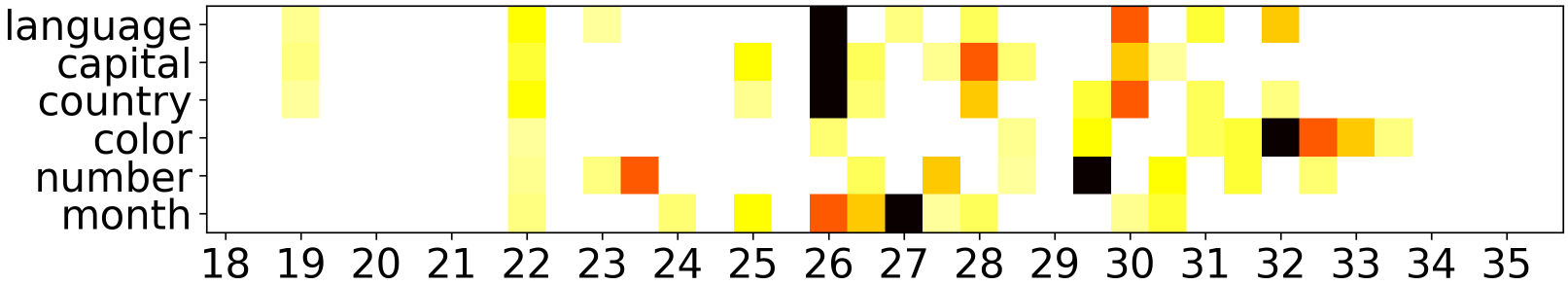}
  \caption{Top10 important "value layers" in GPT2.}
\vspace{-10pt}
\end{figure}

\begin{figure}[htb]
  \centering
    \includegraphics[width=0.98\columnwidth]{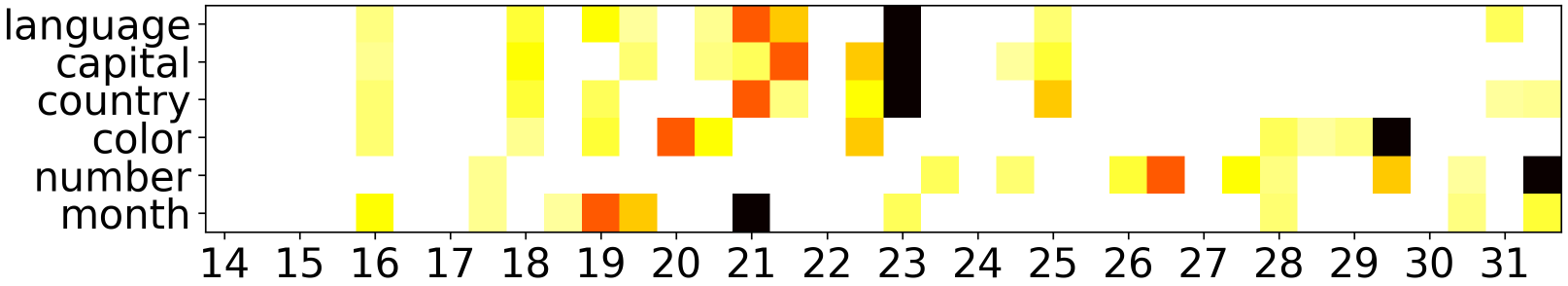}
  \caption{Top10 important "value layers" in Llama.}
\vspace{-10pt}
\end{figure}

\paragraph{Head-level knowledge storage.} We compute the importance score of each head (detailed in Appendix B) and find that many heads have ability to store similar knowledge. In GPT2, $a_{30}^{6}$ (30th layer 6th head), $a_{26}^{17}, a_{32}^{11}, a_{25}^{13}$ and $a_{22}^{17}$ rank top8 for language, capital and country. Similarly, $a_{23}^{12}, a_{19}^{31}, a_{31}^{25}$, and $ a_{25}^{25}$ rank top5 for these knowledge in Llama. 

To evaluate how much knowledge the top heads store, we intervene the top 1\% heads (top7 in GPT2 and top10 in Llama) by setting the heads' parameters to zero. Intervening each knowledge's heads result in a MRR/probability decrease of 44.5\%/53.3\% in GPT2, and 32.8\%/48.2\% in Llama (shown in Appendix B). But semantic-unrelated knowledge only reduce 7.1\%/9.5\% in GPT2 and 3.8\%/8.7\% in Llama. Therefore, the identified "knowledge heads" contain much semantic-related knowledge. 

\begin{table}[htb]
  \centering
  \scalebox{0.91}{\begin{tabular}{ccccccc}
    \toprule
        & lang & capi & cnty & col & num  & mon \\
    \midrule
    (attn) all & 6.7 & 5.5 & 6.9 & 4.0 & 2.4  & 5.4 \\
    positive & 30.5 & 29.4 & 30.5 & 32.0 & 24.8  & 29.2 \\
    top100 & 3.5 & 3.0 & 3.6 & 2.8 & 2.0 & 2.8 \\
    top200 & 5.0 & 4.4 & 5.0 & 4.1 & 3.0 & 4.1 \\
    \midrule
    (FFN) all & 2.5 & 5.1 & 3.6 & 4.9 & 6.5  & 2.8 \\
    positive & 77.4 & 77.6 & 69.8 & 90.6 & 71.6  & 69.1 \\
    top100 & 6.4 & 6.3 & 5.5 & 6.1 & 6.6  & 7.0 \\
    top200 & 8.2 & 8.0 & 7.0 & 8.0 & 8.5  & 8.5 \\
    \bottomrule
  \end{tabular}}
  \caption{Importance of top neurons in attention (first block) and FFN (second block) layers in Llama-7B.}
\vspace{-10pt}
\end{table}

\paragraph{Neuron-level knowledge storage.} For attention and FFN layers in Llama, we compute the sum of importance score for all neurons, all positive neurons (score larger than 0), top100 neurons, and top200 neurons, as illustrated in Table 5. Similar results of GPT2 is shown in Appendix C.

In both models, the sum score of top200 neurons in attention layers and top100 neurons in FFN layers are similar to that of all neurons. Additionally, we intervene the top neurons to evaluate how much final predictions are affected, detailed in Appendix C. When intervening the top200 attention neurons and top100 FFN neurons for each sentence, the MRR and probability decreases 96.3\%/99.2\% in GPT2, and 96.9\%/99.6\% in Llama. In comparison, randomly intervening the same number of neurons only decreases 0.22\%/0.14\%. Hence, even though there are many neurons contribute to the final prediction, intervening a few neurons (300) affects the final prediction much. 
This conclusion holds significance for future studies delving into neuron-level knowledge editing.

\begin{table}[htb]
    \centering
    \scalebox{0.925}{\begin{tabular}{cc}
        \toprule
         & top10 query layers for FFN neurons\\
        \midrule
        lang & $a_{26}, a_{22}, a_{19}, f_{26}, a_{17}, f_{25}, f_{27}, f_{23}, a_{25}, a_{23}$\\
        capi & $a_{26},a_{22},a_{19},a_{17},f_{25},a_{23},a_{18},f_{26},f_{21},a_{28}$ \\  
        cnty & $a_{26},a_{22},a_{17},a_{19},a_{23},f_{18},a_{20},a_{18},f_{21},f_{25}$ \\
        col & $f_{26}, f_{29}, a_{26}, f_{28}, f_{25}, a_{22}, f_{27}, a_{17}, a_{24}, f_{23}$ \\
        num & $f_{26},f_{23},f_{27},a_{22},f_{25},a_{17},f_{19},f_{21},a_{23},f_{0}$ \\
        mon & $a_{17},a_{22},f_{23},a_{26},f_{26},f_{24},a_{19},a_{20},f_{20},f_{21}$ \\
        \midrule
        lang & $f_{21}, a_{16}, a_{19}, f_{18}, a_{18}, a_{21}, a_{17}, f_{30}, f_{19}, a_{14}$ \\
        capi & $a_{18},a_{16},f_{23},a_{19},f_{17},a_{14},f_{22},a_{21},f_{26},f_{19}$\\  
        cnty & $a_{16},a_{18},a_{21},a_{19},f_{21},a_{14},f_{19},a_{17},a_{31},f_{20}$\\  
        col & $f_{20}, f_{21}, a_{15}, a_{17}, a_{18}, a_{20}, f_{19}, f_{22}, f_{17}, a_{16}$ \\
        num & $f_{19},f_{21},f_{22},f_{16},a_{18},a_{22},a_{24},f_{14},a_{12},a_{25}$ \\
        mon & $a_{16},a_{19},f_{18},a_{21},a_{17},a_{14},f_{29}, f_{19}, f_{17}, a_{18}$ \\
        \bottomrule
    \end{tabular}}
    \caption{Top10 query layers for top100 FFN neurons in GPT2 (first block) and Llama (second block).}
\vspace{-10pt}
\end{table}

\begin{figure}[htb]
  \centering
    \includegraphics[width=0.98\columnwidth]{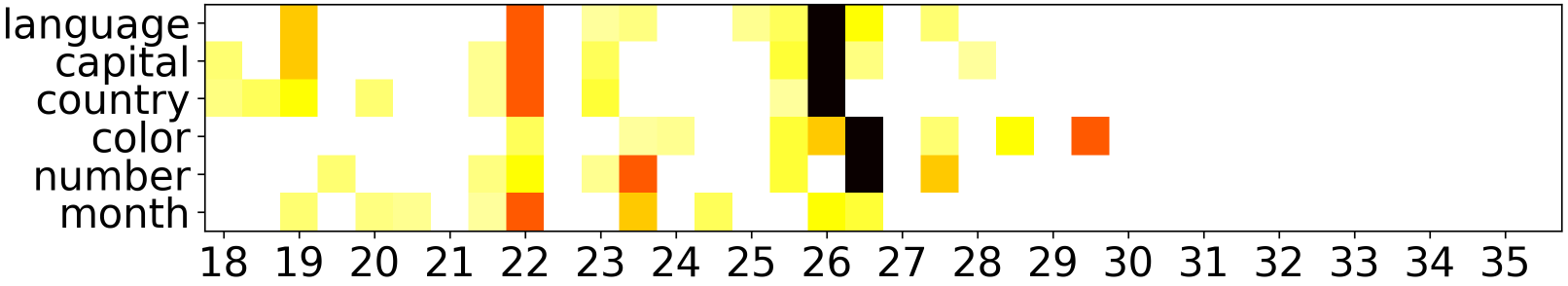}
  \caption{Top10 important "query layers" in GPT2.}
\vspace{-10pt}
\end{figure}

\begin{figure}[htb]
  \centering
    \includegraphics[width=0.98\columnwidth]{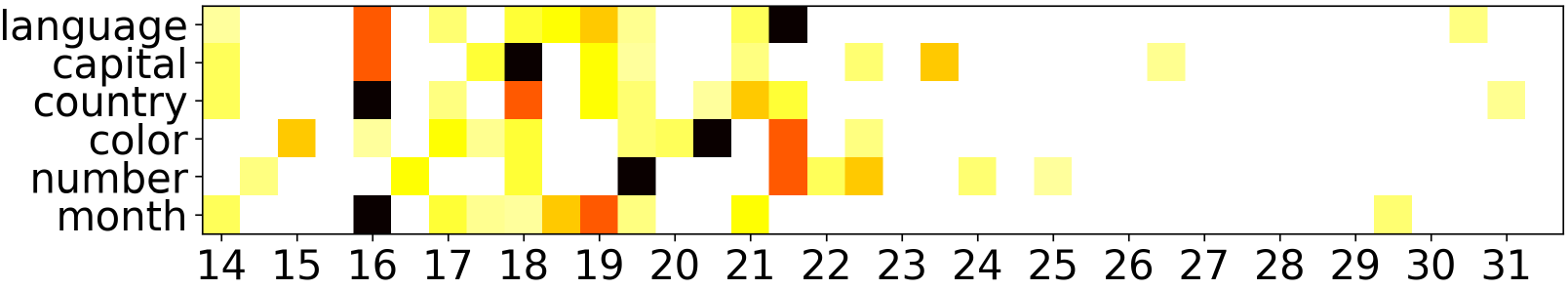}
  \caption{Top10 important "query layers" in Llama.}
\vspace{-10pt}
\end{figure}

\paragraph{Important query layers for FFN value neurons.} The "value" FFN neurons are activated by last position's residual stream. We evaluate which "query layers" activate the top100 FFN neurons, shown in Table 6. We also illustrate the importance of top10 important "query layers" in Figure 6 (GPT2) and Figure 7 (Llama). The medium-deep attention layers play large rules. Compared Figure 6-7 with Figure 4-5, we find that several attention "query layers" also contribute to final predictions (e.g. $a_{19}, a_{22}, a_{26}$ in GPT2 and $a_{16}, a_{18}, a_{19}, a_{21}$ in Llama for country/capital/language). These medium-deep attention layers' neurons are very important, working as both "value" and "query".

\begin{table}[htb]
  \centering
  \scalebox{0.90}{\begin{tabular}{ccccccc}
    \toprule
        & lang & capi & cnty & col & num & mon \\
    \midrule
    G & 91/96 & 98/99 & 89/94 & 95/97 & 96/96  & 83/88 \\
    L & 78/92 & 84/93 & 91/98 & 85/93 & 90/96  & 94/98 \\
    \bottomrule
  \end{tabular}}
  \caption{MRR/probability decrease (\%) when intervening 1,000 query neurons in GPT2 (G) and Llama (L).}
\vspace{-10pt}
\end{table}

\paragraph{Important query neurons for attention value neurons.} We compute the important query layers activating the top200 "value" attention neurons. finding the shallow and medium FFN layers play main roles (detailed in Appendix D). To identify the important query FFN neurons, we weighted sum the inner product between each attention neuron's subkey and each FFN neuron on every position's residual stream, as query FFN neurons' scores. When intervening top1000 shallow neurons for each sentence, both MRR and probability drops very much (92\%/95\% in GPT2 and 87\%/95\% in Llama), shown in Table 7. In comparison, randomly intervening 1,000 neurons only result in a decrease of 0.8\%/1.1\%. Hence, our method can locate the important query neurons in these layers.

\begin{figure}[htb]
  \centering
  \includegraphics[width=0.92\columnwidth]{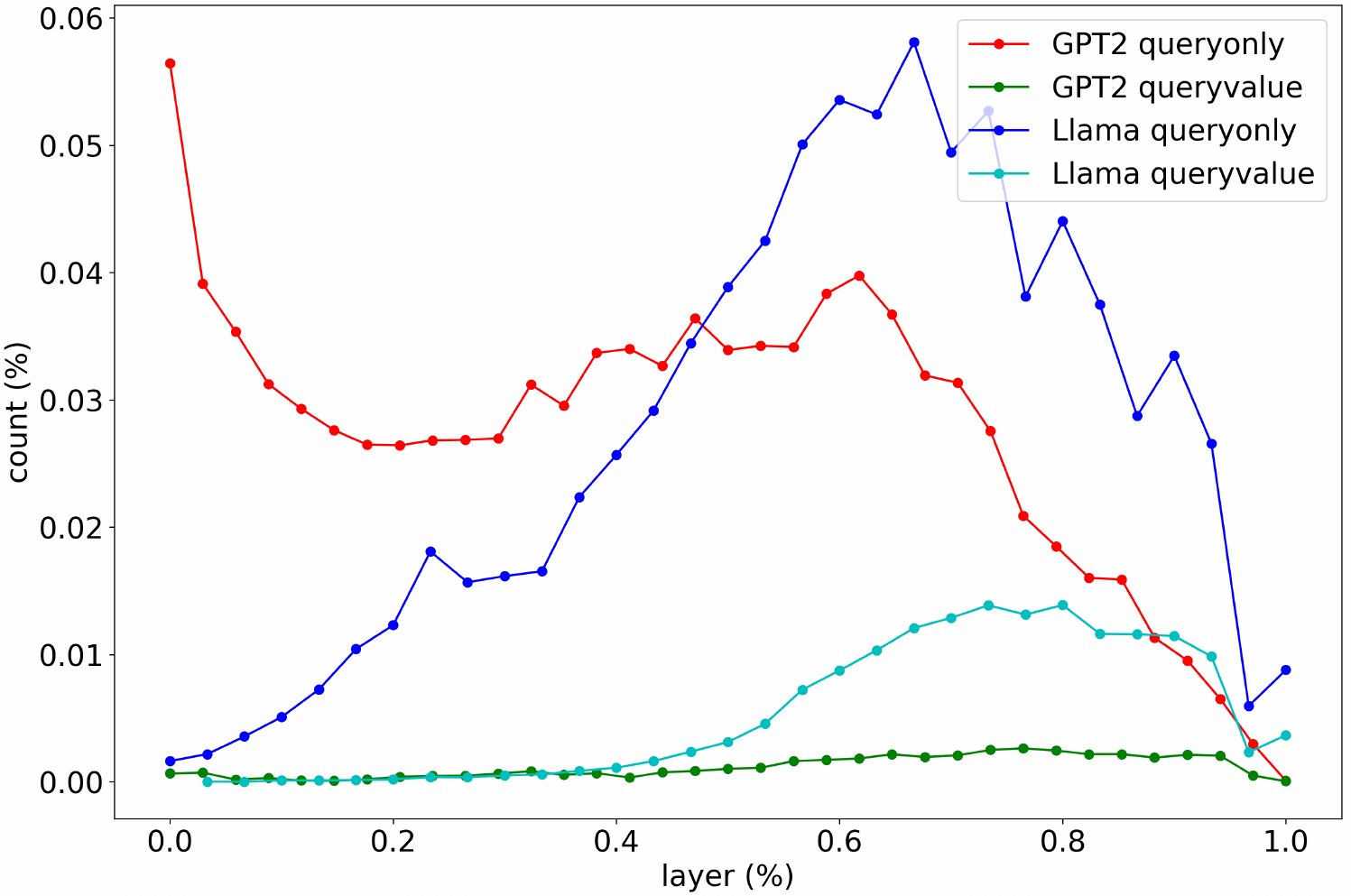}
  \caption{Query neuron distribution in GPT2 and Llama.}
\vspace{-10pt}
\end{figure}

Then we count the number of queryvalue (both in top1000 query and top1000 value) and queryonly (only in top1000 query) FFN neurons, shown in Figure 8. In both models, the number of queryonly neurons, which is much larger than that of queryvalue neurons, starts to drop at 60\% layer. This observation indicates that the shallow and medium FFN neurons are important for activating the attention "value neurons". A difference is that the very shallow FFN layers play large roles in GPT2, and we defer this exploration to future research. Overall, our analysis learns the information flow at neuron level: features in shallow/medium FFN neurons are extracted, then activate the deep attention and FFN neurons related to final predictions.

\begin{table}[htb]
  \centering
  \scalebox{0.92}{\begin{tabular}{ccccccc}
    \toprule
        & lang & capi & cnty & col & num & mon \\
    \midrule
    G-query & 5 & 51 & 26 & 17 & 104 & 81 \\
    G-value & 48 & 71 & 64 & 57 & 121 & 137 \\
    L-query & 1 & 1 & 1 & 9 & 39 & 44 \\
    L-value & 13 & 18 & 21 & 23 & 84 & 95 \\
    \bottomrule
  \end{tabular}}
  \caption{Shared neurons in GPT2 (G) and Llama (L).}
\vspace{-10pt}
\end{table}

\paragraph{Shared Value and query neurons in each knowledge.} We compute how many "shared" query neurons  and value neurons rank top300 in more than 50\% sentences in each knowledge, shown in Table 8. On average, there are 27.6\% shared value neurons in GPT2 and 14.1\% in Llama. Query neurons, with 15.7\% shared neurons in GPT2 and 5.2\% in Llama, exhibit a more dispersed distribution than value neurons. To explore the neurons' interpretability, we project them into vocabulary space. We find most value neurons (first block in Table 9) are related to predicted tokens. However, we do not observe much interpretability in query neurons. We only find a few query neurons (second block in Table 9) related to the final words. Hence, to explore the interpretability of query neurons may be a valuable direction in future works.

\begin{table}[htb]
    \centering
    \scalebox{0.90}{\begin{tabular}{p{1.5cm}p{6cm}}
        \toprule
        neuron & top10 tokens in vocabulary space\\
        \midrule
        $f_{29}$-3771 (GPT2,v) & Chile,Nicaragua,Finland,Ireland,Belarus, Norway,Slovakia,Latvia,Australia\\
        $a_{23}^{12}$-70 (Llama,v) & German,Greek,Netherlands,Dutch, Germany,Greece,Holland,Norwegian\\
        \hline
        $f_0$-2947 (GPT2,q) & Lion, Bull, Liver, riot, Gladiator, \textbf{Red}, uct, Ant, Les\\
        $f_7$-8744 (Llama,q) & \textbf{Belgium},Ireland,Vienna,Czech,Kas, Kansas,Netherlands,Iowa,wings,Spanish\\
        \bottomrule
    \end{tabular}}
    \caption{Interpretable neurons in vocabulary space.}
\vspace{-10pt}
\end{table}

\section{Conclusion}
In this study, we propose a method based on log probability increase to identify the important "value neurons". We also develop a method based on inner products to locate the "query neurons" activating these "value neurons". Our method and analysis on six types of knowledge are helpful for exploring and understanding the mechanism of LLMs.

\section{Limitations}
The first limitation of our study is that it focuses on six specific types of knowledge, while other types of knowledge are also important. Secondly, our experiments are conducted using GPT2-large and Llama-7B models. It is essential to compare the similarities and differences in knowledge storage across different models. Lastly, our study employs static methods for neuron-level knowledge attribution. Although our experiments demonstrate the correctness and robustness of our designed method, it is also important to compare static methods with other attribution methods, such as causal mediation analysis and gradient-based methods. We plan to explore these areas in future work.

A potential risk of our work is that people can utilize our method to identify important neurons and edit them to change the models' outputs. For instance, if they identify the toxicity neurons and gender bias neurons and increase these neurons' coefficient scores, the model will be more likely to generate toxicity and gender bias words. But this potential risk depends on how people utilize our method. Our method can be utilized for reducing hallucinations, toxicity, and bias in LLMs by identifying and intervening/editing these neurons.

\section{Acknowledgements}
We thank Kailai Yang, Zhiwei Liu, and John McNaught for helpful feedbacks and constructive suggestions. This work is supported by the project JPNP20006 from New Energy and Industrial Technology Development Organization (NEDO). This work is supported by the computational shared facility and the studentship from the Department of Computer Science at the University of Manchester. 

\bibliography{custom}
\bibliographystyle{acl_natbib}

\clearpage
\appendix
\section{Neuron Distribution in GPT2}
The neuron distribution of GPT2-large is similar to Llama-7B, which is shown in Figure 9. Also, the curve of importance score in GPT2-large is similar to that in Llama-7B, illustrated in Figure 10.

\begin{figure}[htb]
  \centering
  \includegraphics[width=0.92\columnwidth]{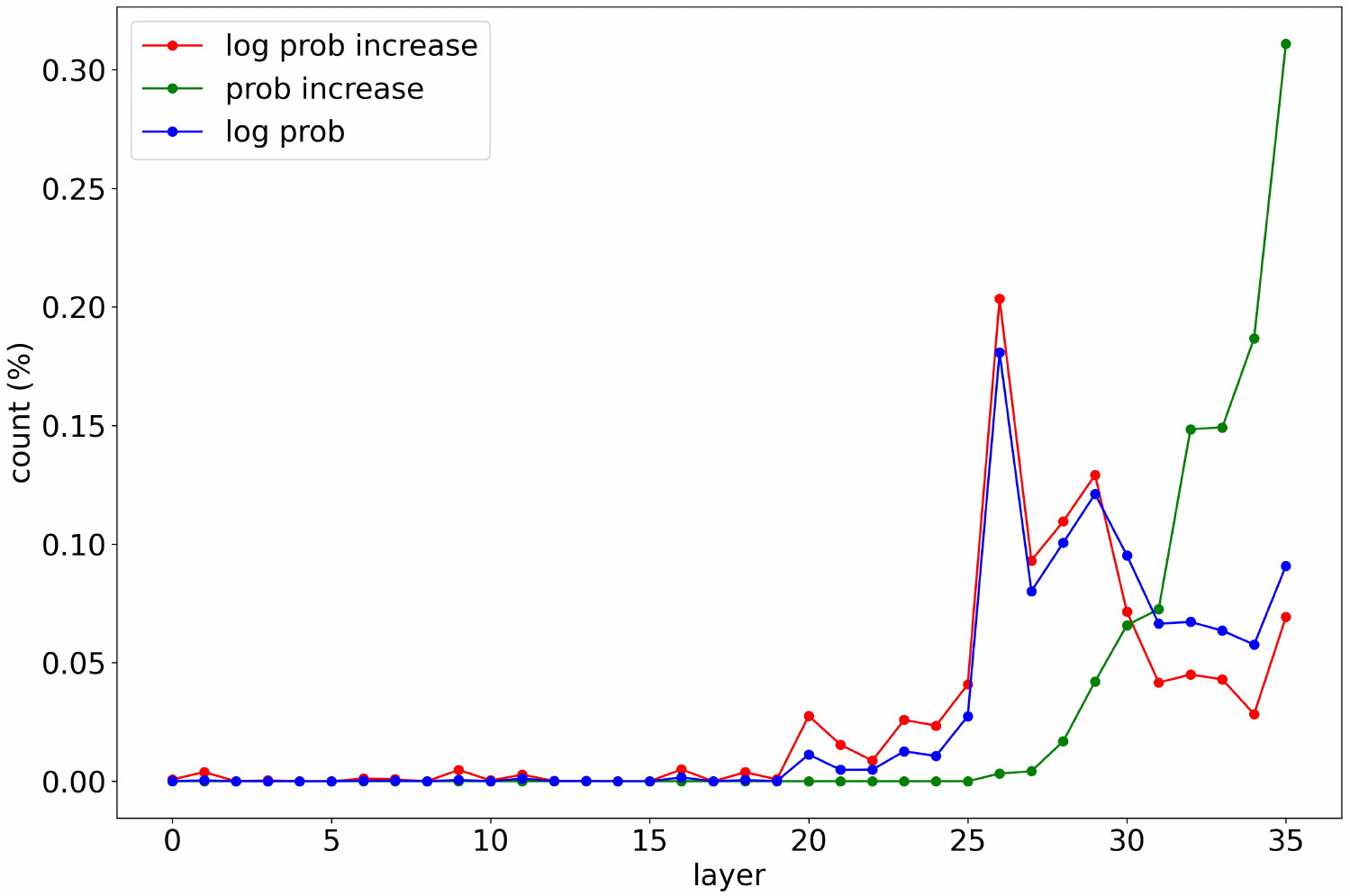}
  \caption{Neuron distribution on all layers in GPT2.}
\vspace{-10pt}
\end{figure}

\begin{figure}[htb]
	\centering
	\subfigure{
		\begin{minipage}{0.48\textwidth}
			\includegraphics[width=0.47\textwidth]{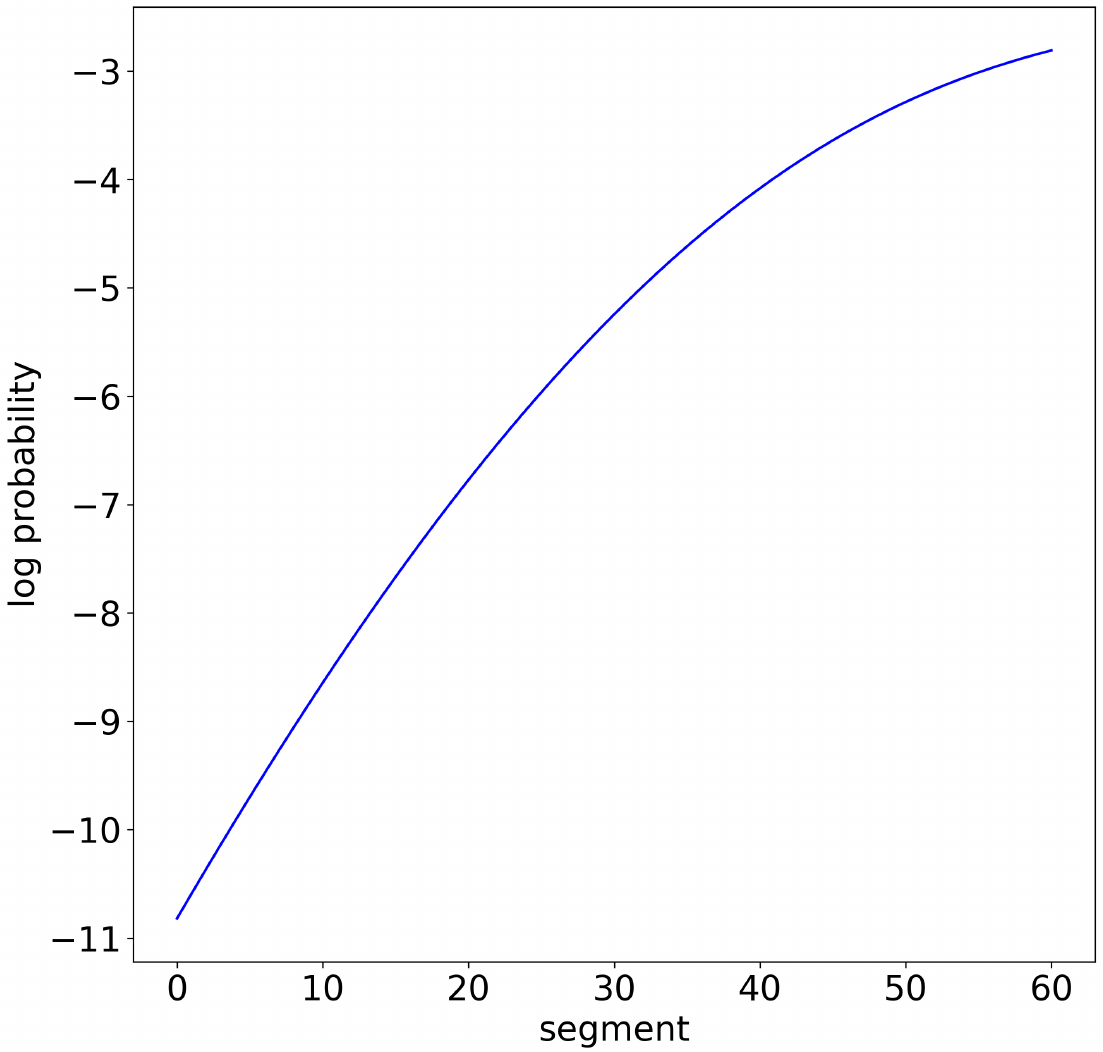}
                \includegraphics[width=0.47\textwidth]{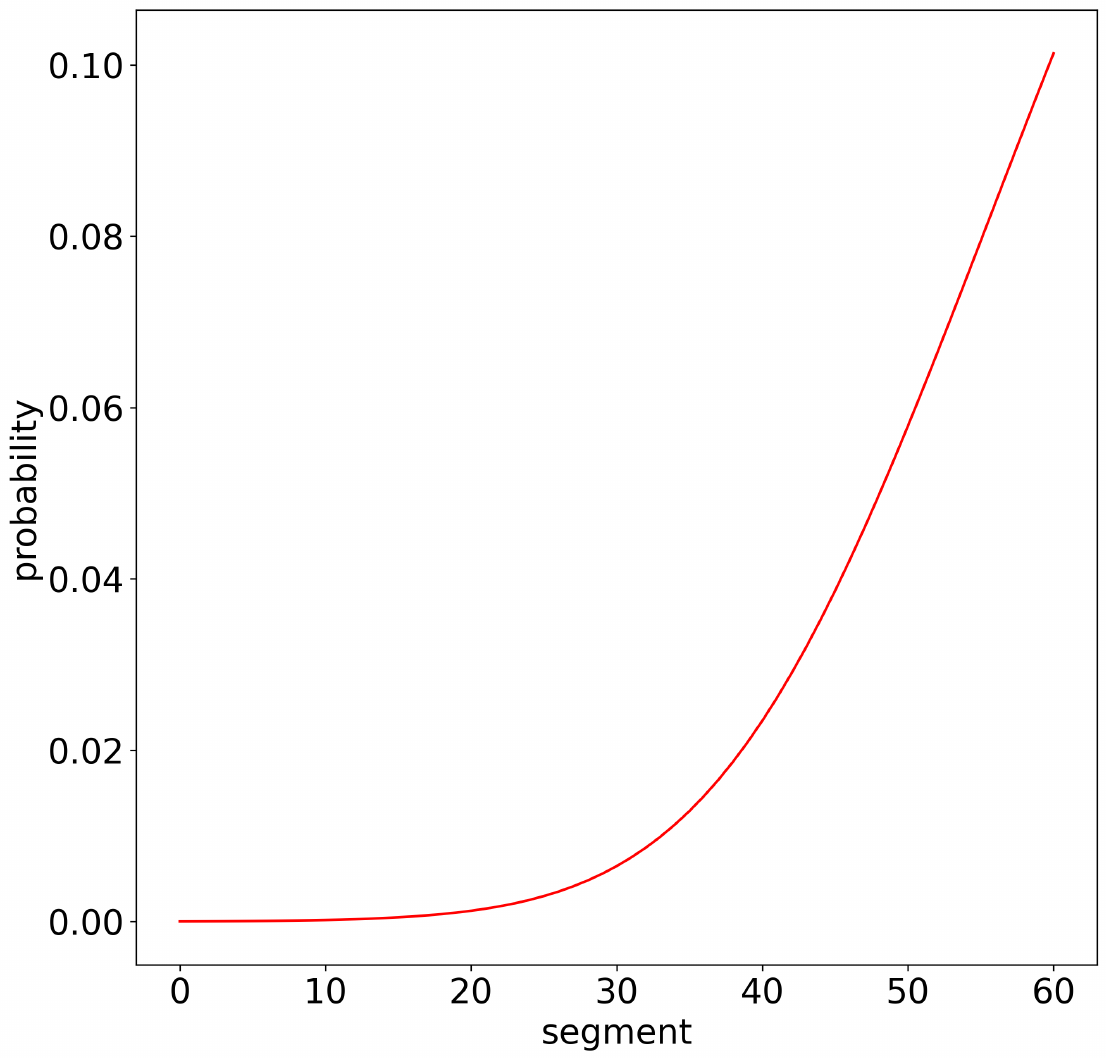}
		\end{minipage}
	}
	\caption{Curves of log probability increase (left) and probability increase (right) on GPT2.}
\vspace{-10pt}
\end{figure}

\section{Head-Level Storage in GPT2/Llama}
The top10 heads are shown in Table 10, where $a_l^j$ is short for the $jth$ head in $lth$ attention layer. Knowledge with similar semantics is stored in the same heads (e.g. $a_{30}^6$ in GPT2 and $a_{23}^{12}$ in Llama).

\begin{table}[htb]
    \centering
    \scalebox{0.925}{\begin{tabular}{cc}
        \toprule
        type & top10 heads \\
        \midrule
        lang & $a_{30}^{6}, a_{26}^{17}, a_{26}^{7}, a_{32}^{11}, a_{19}^{0}, a_{31}^{9}, a_{25}^{13}, a_{22}^{17}, a_{28}^{13}, a_{29}^{2}$\\
        capi & $a_{26}^{7}, a_{30}^{6}, a_{26}^{17}, a_{22}^{17}, a_{25}^{13}, a_{28}^{13}, a_{19}^{0}, a_{19}^{10}, a_{29}^{2}, a_{32}^{11}$ \\  
        cnty & $a_{26}^{7}, a_{30}^{6}, a_{22}^{17}, a_{28}^{13}, a_{26}^{17}, a_{32}^{11}, a_{19}^{0}, a_{25}^{13}, a_{31}^{9}, a_{19}^{10}$ \\  
        col & $a_{33}^{5}, a_{34}^{1}, a_{26}^{7}, a_{24}^{19}, a_{23}^{18}, a_{32}^{13}, a_{30}^{1}, a_{22}^{8}, a_{32}^{14}, a_{28}^{2}$ \\
        num & $a_{22}^{18}, a_{17}^{3}, a_{23}^{8}, a_{19}^{2}, a_{30}^{3}, a_{25}^{19}, a_{20}^{3}, a_{30}^{0}, a_{12}^{2}, a_{25}^{3}$ \\
        mon & $a_{27}^{2}, a_{26}^{7}, a_{25}^{11}, a_{19}^{10}, a_{30}^{2}, a_{28}^{4}, a_{23}^{18}, a_{17}^{17}, a_{33}^{1}, a_{17}^{3}$ \\
        \midrule
        lang & $a_{23}^{12}, a_{19}^{31}, a_{31}^{25}, a_{25}^{25}, a_{16}^{5}, a_{18}^{1}, a_{21}^{9}, a_{29}^{22}, a_{21}^{17}, a_{18}^{23}$ \\
        capi & $a_{23}^{12}, a_{29}^{22}, a_{25}^{25}, a_{31}^{25}, a_{19}^{31}, a_{18}^{1}, a_{16}^{15}, a_{16}^{5}, a_{21}^{9}, a_{18}^{23}$ \\  
        cnty & $a_{23}^{12}, a_{19}^{31}, a_{25}^{25}, a_{21}^{9}, a_{31}^{25}, a_{16}^{15}, a_{18}^{1}, a_{16}^{5}, a_{29}^{22}, a_{28}^{19}$ \\  
        col & $a_{29}^{22}, a_{28}^{19}, a_{20}^{27}, a_{16}^{15}, a_{17}^{27}, a_{28}^{21}, a_{25}^{14}, a_{18}^{28}, a_{24}^{1}, a_{14}^{3}$ \\
        num & $a_{28}^{19}, a_{26}^{24}, a_{23}^{10}, a_{30}^{13}, a_{21}^{29}, a_{13}^{24}, a_{18}^{24}, a_{29}^{22}, a_{17}^{23}, a_{19}^{1}$ \\
        mon & $a_{21}^{10}, a_{16}^{0}, a_{21}^{22}, a_{23}^{18}, a_{28}^{16}, a_{19}^{20}, a_{31}^{6}, a_{19}^{1}, a_{14}^{3}, a_{20}^{13}$ \\
        \bottomrule
    \end{tabular}}
    \caption{Top10 important heads in GPT2 (first block) and Llama (second block).}
\vspace{-10pt}
\end{table}

The MRR decrease (\%)/probability decrease (\%) when intervening the top 1\% heads for each knowledge is shown in Table 11. When intervening the top 1\% heads for each knowledge, similar knowledge (language, capital and country) is affected a lot, while other knowledge (month, color, number) is not affected much.

\begin{table}[htb]
  \centering
  \scalebox{0.85}{\begin{tabular}{ccccccc}
    \toprule
        & lang & capi & cnty & mon & col & num \\
    \midrule
    lang & \textbf{44/51} & \textbf{33/52} & \textbf{38/56} & 3/0 & 6/5 & 1/2 \\
    capi & \textbf{32/38} & \textbf{42/53} & \textbf{39/54} & 15/11 & 18/12 & 0/1 \\
    cnty & \textbf{39/44} & \textbf{40/54} & \textbf{44/60} & 3/0 & 10/5 & 2/3 \\
    mon & 19/24 & 14/19 & 13/19 & \textbf{55/63} & 24/23 & 5/4 \\
    col & 6/6 & 1/0 & 2/4 & 13/12 & \textbf{49/59} & 5/7 \\
    num & 11/14 & 2/10 & 7/11 & 17/21 & 19/16 & \textbf{33/34} \\
    \midrule
    lang & \textbf{24/42} & \textbf{17/35} & \textbf{13/33} & 0/0 & 6/15 & 1/1 \\
    capi & \textbf{38/58} & \textbf{28/50} & \textbf{22/53} & 1/0 & 7/16 & 0/0 \\
    cnty & \textbf{42/61} & \textbf{31/54} & \textbf{28/58} & 0/0 & 10/21 & 2/6 \\
    mon & 7/14 & 4/11 & 7/13 & \textbf{51/66} & 8/13 & 3/8 \\
    col & 3/12 & 6/16 & 6/14 & 13/25 & \textbf{33/42} & 1/10 \\
    num & 0/0 & 1/4 & 1/2 & 2/9 & 3/13 & \textbf{33/31} \\
    \bottomrule
  \end{tabular}}
  \caption{MRR decrease (\%)/probability decrease (\%) in GPT2 (first block) and Llama (second block) when intervening top 1\% heads for different knowledge.}
\vspace{-10pt}
\end{table}

\section{Neuron-Level Storage in GPT2/Llama}
The sum score of top neurons and all neurons in GPT2 are shown in Table 12. The sum importance score of top200 attention neurons and top100 FFN neurons are similar to those of all neurons.

\begin{table}[htb]
  \centering
  \scalebox{0.92}{\begin{tabular}{ccccccc}
    \toprule
        & lang & capi & cnty & col & num  & mon \\
    \midrule
    all & 7.3 & 6.7 & 6.6 & 3.6 & 2.3  & 4.3 \\
    positive & 28.8 & 27.4 & 27.5 & 24.7 & 18.2  & 23.3 \\
    top100 & 4.0 & 3.7 & 3.5 & 2.7 & 1.6  & 2.5 \\
    top200 & 5.7 & 5.3 & 5.1 & 4.0 & 2.5  & 3.7 \\
    \midrule
    all & 0.0 & 1.9 & 1.4 & 2.5 & 4.0  & 1.9 \\
    positive & 70.1 & 69.4 & 69.1 & 72.0 & 62.3  & 66.6 \\
    top100 & 4.2 & 4.6 & 4.3 & 4.3 & 4.1  & 4.7 \\
    top200 & 6.0 & 6.3 & 5.9 & 6.1 & 5.7  & 6.4 \\
    \bottomrule
  \end{tabular}}
  \caption{Imporatnce of top neurons in attention (first block) and FFN (second block) layers in GPT2.}
\vspace{-10pt}
\end{table}

The MRR decrease (\%) / probability decrease (\%) when intervening the top 200 attention neurons and top100 FFN neurons are shown in Table 13. The MRR score and probability score decreases around 91.1\%/98.7\% in GPT2, and 88.4\%/97.1\% in Llama. Therefore, our method can identify the important "value neurons" in both attention and FFN layers.

\begin{table}[htb]
  \centering
  \scalebox{0.90}{\begin{tabular}{ccccccc}
    \toprule
        & lang & capi & cnty & col & num  & mon \\
    \midrule
    G & 96/99 & 96/99 & 97/99 & 97/99 & 96/98  & 96/99 \\
    \midrule
    L & 97/99 & 97/99 & 97/99 & 98/99 & 96/99  & 97/99 \\
    \bottomrule
  \end{tabular}}
  \caption{MRR decrease (\%) and probability decrease (\%) when intervening the top200 attention neurons and top100 FFN neurons in GPT (G) and Llama (L).}
\vspace{-10pt}
\end{table}

\section{Important Query Layers for Attention Neurons in GPT2/Llama}
We evaluate which layers have large inner product with top200 attention neurons, shown in Table 14. For every knowledge, the shallow and medium FFN layers play larger roles than attention layers.

\begin{table}[htb]
    \centering
    \scalebox{0.925}{\begin{tabular}{cc}
        \toprule
         & top10 query layers for attention neurons\\
        \midrule
        lang & $f_0, f_1, a_0, f_2, f_{19}, f_{20}, f_3, f_{17}, f_{18}, f_{21}$ \\
        capi & $f_0, f_1, a_0, f_2, f_3, f_{20}, f_5, f_4, f_{19}, f_{17}$  \\  
        cnty & $f_0, f_1, f_{19}, a_0, f_{18}, f_2, f_3, f_{21}, f_{20}, f_{17}$  \\ 
        col & $f_0, f_1, f_2, f_{23}, f_{20}, f_{21}, f_{22}, f_{24}, a_0, f_3$  \\
        num & $f_0, f_{18}, f_1, f_{19}, f_{22}, f_{16}, f_{21}, f_2, f_{12}, f_{20}$  \\
        mon & $f_0, f_1, f_{19}, f_2, f_9, f_{22}, f_{10}, f_{21}, a_0, f_{18}$ \\
        \midrule
        lang & $f_{20},f_{19},a_{16},f_{16},f_{15},f_{18},f_{14},f_{21},f_{12},f_{21}$ \\
        capi & $f_{20},f_{24},f_{22},f_{23},f_{19},a_{16},a_{23},f_{28},f_{18},f_{25}$ \\  
        cnty & $f_{18},f_{21},f_{19},a_{18},f_{22},a_{14},a_{16},f_{17},a_{21},a_{19}$ \\  
        col & $f_{15},f_{18},f_{20},f_{16},f_{19},f_{13},f_{17},f_{22},f_{24},f_{14}$ \\
        num & $f_{24},f_{17},f_{19},f_{23},f_{22},f_{20},f_{18},f_{2},f_{25},f_{21}$ \\
        mon & $f_{14},a_{16},f_{19},a_{19},f_{18},f_{20},f_{13},f_{17},f_{15},f_{22}$ \\
        \bottomrule
    \end{tabular}}
    \caption{Top10 query layers for top200 attention neurons in GPT2 (first block) and Llama (second block).}
\vspace{-10pt}
\end{table}

\end{document}